\documentclass{article}

\PassOptionsToPackage{numbers, compress}{natbib}

\usepackage[final]{neurips_2020}

\usepackage[utf8]{inputenc} 
\usepackage[T1]{fontenc}    
\usepackage{hyperref}       
\usepackage{url}            
\usepackage{booktabs}       
\usepackage{amsfonts}       
\usepackage{nicefrac}       
\usepackage{microtype}      

\usepackage{graphicx}
\usepackage{subfigure}
\usepackage{multirow}
\usepackage{amssymb}
\usepackage{amsmath}
\usepackage{booktabs}
\usepackage{algorithm}
\usepackage{algorithmic}
\usepackage{color}
\usepackage{wrapfig}

\usepackage[dvipsnames]{xcolor}

\usepackage{array}
\usepackage{stfloats}
\usepackage{url}
\usepackage{ragged2e}
\usepackage{bm}

\title{Domain Contrast for Domain Adaptive Object Detection}

\author{Feng Liu, Xiaoxong Zhang, Fang Wan, Xiangyang Ji, Qixiang Ye \thanks{Qixiang Ye is the corresponding author. This work was supported by the National Natural Science Foundation of China (NSFC) under Grant 61836012, 61671427, and 61771447.} \\
  Electronic, Electrical and Communication Engineering\\
  University of Chinese Academy of Sciences, Beijing, China, 100049.\\
  \texttt{\{qxye\}@ucas.ac.cn} \\
}

\begin{document}

\maketitle

\begin{abstract}
    We present Domain Contrast (DC), a simple yet effective approach inspired by contrastive learning for training domain adaptive detectors. DC is deduced from the error bound minimization perspective of a transferred model, and is implemented with cross-domain contrast loss which is plug-and-play. By minimizing cross-domain contrast loss, DC guarantees the transferability of detectors while naturally alleviating the class imbalance issue in the target domain. DC can be applied at either image level or region level, consistently improving detectors' transferability and discriminability. Extensive experiments on commonly used benchmarks show that DC improves the baseline and state-of-the-art by significant margins, while demonstrating great potential for large domain divergence.
\end{abstract}

\section{Introduction}
\label{sec:intro}
    Modern object detectors~\cite{ren2015faster,liu2016ssd} have achieved unprecedented progress with the rise of convolutional neural networks (CNNs). Although, their practical application to real-world scenarios remains limited for the following two reasons: 1) Supervised learning of detectors for different scenarios requires repeated human effort on data annotation, and 2) offline-trained detectors typically degrade with changes in the scene or camera. Domain adaptive detection, which transfers detectors trained within a label-rich domain ($i.e.$, annotated datasets) to an unlabeled domain ($i.e.$, real-world scenarios), have attracted increasing interests because of their potential to solve these problems~\cite{chen2018domain,saito2019strongweak,zhu2019selective,wang2019domainAttention}.

    Domain adaptive detection is usually explored with unsupervised domain adaptation (UDA) methods. Such methods typically maximize the ``transferability" of models by aligning the feature distributions of domains~\cite{chen2018domain} or taking advantages of adversarial and generative models~\cite{Konstantinos2017ga,Judy2018cycada}. The underlying hypothesis is that accurate feature alignment across domains produces good transferability. 
    
    Recent studies~\cite{Chen2020Transferability} have shown that transferability and discriminability are in fact two sides of a same coin. The transferring process, $e.g.,$ minimizing Maximum Mean Discrepancy (MMD)~\cite{gretton2012mmd}, could deteriorate the discriminability of models and features~\cite{LargeNorm2019}. This would be more severe in the object detection problem, considering the large imbalance of negative and positive instances which need to be classified. Slight degradation of model discriminability could therefore cause a significant increase of false positives in 
    the target domain. 
    
    \begin{figure}[t]
		\centering
		\includegraphics[width = 0.75\linewidth]{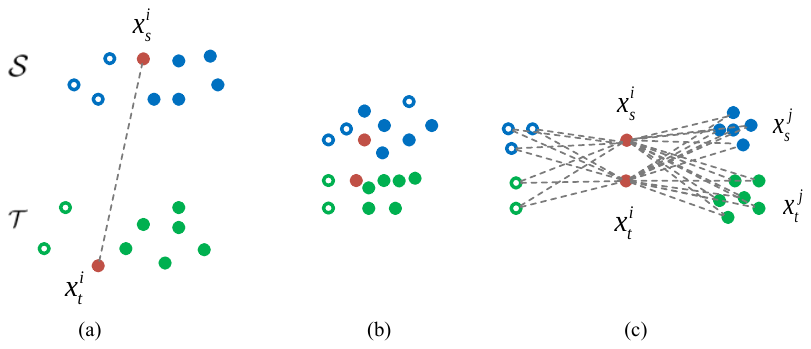}
		\caption{Illustration of Domain Contrast. (a) The reduction of domain divergence (distance between $x_s^i$ and $x_t^i$) towards aligning feature distributions of the source domain $\mathcal{S}$ and the target domain $\mathcal{T}$. (b) The reduction of the domain divergence could reduce the discriminability of models, $i.e.$, samples from different categories are mixed together. (c) Domain contrast simultaneously reduces domain divergence by minimizing the cross-domain distance while preserving model discriminability by maximizing the distances between samples from different categories. }
		\label{fig:dc}
	\end{figure}

	In this paper, we derive the Domain Contrast (DC) method from the perspective of error bound minimization when transferring detectors from a source to a target domain. Minimizing error bound is converted to optimize DC loss, which defines the cross-domain similarity and inter-class distance for each min-batch of samples, Fig.\ \ref{fig:dc}.
	DC guarantees the transferability of detectors by maximizing the $cosine$ similarity between each sample with its cross-domain counterparts. This also preserves the discriminability of transferred detectors by minimizing the similarity between samples from different categories in a mini-batch.
	
	To define DC, training sample images are translated from the source/target to the target/source domain via a CycleGAN method~\cite{CycleGAN2017}. With translated images, a detector is trained using annotated samples in the source domain and fine-tuned using DC loss from the source to the target domain. In turn, the detector is fine-tuned using the DC loss from the target to the source domain. DC loss is applied at image- and region-level, consistently improving the transferability and discriminability of detectors.
	
	The contributions of this work are summarized as follows: 
	\begin{itemize}
		\item A simple-yet-effective domain contrast (DC) method for transfer learning, where the DC loss is derived from the perspective of error bound minimization. Minimization of DC loss drives feature alignment ($i.e.$, transferability) across domains while preserving model discriminability in the target domain. 
		
		\item A domain adaptive detector considering the class imbalance issue. The detector leverages DC loss at image level and region level to handle the detector transfer problem, achieving state-of-the-art performance on benchmarks with large domain divergence.
		
	\end{itemize}

\section{Related Work}
\label{sec:related}

  	\textbf{Unsupervised Domain Adaptation (UDA).}
	UDA aims to minimize the classification error when transferring the model trained within a source domain to an unlabelled target domain. UDA has been extensively explored in a number of computer vision specializations, including image classification~\cite{li2017lowrank,motiian2017unified,panareda2017open,ganin2014GRL,Long2015DAN}, image recognition~\cite{deng2018spgan,zhong2019invariance}, and object detection~\cite{chen2018domain,saito2019strongweak,zhu2019selective,kim2019diversify}. 

	One line of UDA methods devoted to align feature distributions of source and target domains by minimizing the domain divergence~\cite{Long2015DAN,ganin2014GRL}. For example, Maximum Mean Discrepancy (MMD)~\cite{gretton2012mmd} was proposed as a domain distance metric to minimize the domain divergence in the Reproducing Kernel Hilbert Space (RKHS)~\cite{Long2015DAN,long2016unsupervised}. Progressive domain distance minimization~\cite{chen2019progressive} was achieved by mining pseudo labels to fine-tune the model. 
	The other line of methods~\cite{Konstantinos2017ga,Judy2018cycada} attempted to reduce the domain divergence by taking advantages of adversarial and generative models which confuse domains while aligning feature distributions. The CyCADA method~\cite{Judy2018cycada} transferred samples across domains at both pixel- and feature-levels. Domain confusion loss~\cite{ganin2014GRL} was designed to learn domain-invariant features. The descrepancy-based method~\cite{Saito2018discrepancy} aligned distributions of source and target domains by learning features to minimize classifiers' output discrepancy.

	\textbf{Domain Adaptive Detection.}
	Early studies largely followed domain adaptive classification to align the features of source and target domains. DA-Faster R-CNN~\cite{chen2018domain} pioneered these works by minimizing the discrepancy among two domains by exploring both image- and instance-level domain classifier in an adversarial manner. Mean Teacher with object relations~\cite{cai2019meanteacher} was applied for object distribution alignment, while integrating object relations with the measure of consistency cost between teacher and student modules. The Diversify and Match (DM) approach~\cite{kim2019diversify} generated various distinctive shifted domains from the source domain and aligned the distribution of the labeled data and encouraged features to be indistinguishable among the domains. Strong-and-Weak~\cite{saito2019strongweak} method pursue weak alignment of image-level features and strong alignment of region-level features.
	
	Despite progress, the essential difference between domain adaptive detection with domain adaptive classification is unfortunately ignored. 
	On the one hand, source and target domains have distinct scene layouts and object combinations. Therefore, aligning the entire distributions of source and target images is implausible. On the other hand, object detectors face the serious class imbalance issue. Preserving model discriminability during domain adaptive detection is more important than that in image classification~\cite{zhu2019selective}.

	%
	To preserve the discriminability, the Selective Cross-Domain approach~\cite{zhu2019selective} attempted aligning discriminative regions, namely those that are directly related to detection. 
	The Hierarchical Transferability Calibration Network harmonized transferability and discriminability for cross-domain detection~\cite{Chen2020Transferability}. Nevertheless, these approaches used complex adversarial training and/or sample interpolation which hinders deployment and therefore practicability. 
	
\section{The Proposed Approach}
\label{gen_inst}
    Let $\mathcal{S}$ and $\mathcal{T}$ respectively denote a source and a target domain.
    The corresponding samples of $\mathcal{S}$ and $\mathcal{T}$ are denoted as $\{x^i_s\}_{i=1}^{N}$ and $\{x^i_t\}_{i=1}^{N}$.  $f_\mathcal{S}:\mathcal{X} \rightarrow \{0, 1\}$ and $f_\mathcal{T}:\mathcal{X} \rightarrow \{0, 1\}$ denote functions which map the input samples $\mathcal{X}$ to a binary label space. 
    $f_\mathcal{T}(x_s^i) = f_\mathcal{T}(x_t^i)$ and $f_\mathcal{S}(x_s^i) = f_\mathcal{S}(x_t^i)$ mean that the sample labels are consistent regardless of the domains. Domain adaptation targets at transferring a model ($i.e.$, the detector) optimized for $f_\mathcal{S}$ to $\mathcal{T}$ towards optimizing $f_\mathcal{T}$.
    
    
\subsection{Domain Contrast}
     \textbf{Error Bound Minimization}. The source and target domains share an identical label space, but violate the $i.i.d.$ assumption as they are sampled from different data distributions. Given a model hypothesis $h\in \mathcal{H}$, the expected error~\cite{DShai2010bound} within the target domain are bounded as 
    \begin{equation}
    \label{eq:bound}
    \begin{split}
        \mathcal{R}_\mathcal{T}(h, f_{\mathcal{T}}) 
        &\leq \mathcal{R}_\mathcal{T}(h, f_\mathcal{S}) + |\mathcal{R}_\mathcal{T}(h, f_{\mathcal{T}}) - \mathcal{R}_\mathcal{T}(h, f_{\mathcal{S}})|, \\
    \end{split}
    \end{equation}
    where $\mathcal{R}_\mathcal{T}(h, f_{\mathcal{T}})$ and $\mathcal{R}_\mathcal{T}(h, f_\mathcal{S})$  respectively denote the empirical error of hypothesis $h$ about labels in the target and source domains. 
    To minimize the error bound defined by Eq.\ \ref{eq:bound} is to minimize $|\mathcal{R}_\mathcal{T}(h, f_{\mathcal{T}}) - \mathcal{R}_\mathcal{T}(h, f_{\mathcal{S}})|$ and  $\mathcal{R}_\mathcal{T}(h, f_\mathcal{T})$, which aligns the two domains while preserving the discriminability of the trained model in the target domain. 
    
    In the context of CNN, with the binary cross entropy loss, we have
    \begin{equation}
    \label{eq:risk1}
    \begin{split}
        \mathcal{R}_\mathcal{T}(h, f_\mathcal{S}) 
        &= \frac{1}{N} \sum_{i}{\bigg( -f_\mathcal{S}(x_t^i)\log(h(x_t^i)) - (1 - f_\mathcal{S}(x_t^i))\log(1 - h(x_t^i)) \bigg)},
    \end{split}
    \end{equation}
    and
    \begin{equation}
    \label{eq:risk2}
    \begin{split}
        \mathcal{R}_\mathcal{T}(h, f_\mathcal{T}) 
        &= \frac{1}{N} \sum_{i}{\bigg( -f_\mathcal{T}(x_t^i)\log(h(x_t^i)) - (1 - f_\mathcal{T}(x_t^i))\log(1 - h(x_t^i)) \bigg)},
    \end{split}
    \end{equation}
    where $N$ denotes the number of samples. Subtracting Eq.\ \ref{eq:risk2} from Eq.\ \ref{eq:risk1}, we have
    \begin{equation}
    \label{eq:bound1}
    \begin{split}
        |\mathcal{R}_\mathcal{T}(h, f_\mathcal{T}) - \mathcal{R}_\mathcal{T}(h, f_\mathcal{S})| 
        &= \bigg|\frac{1}{N} \sum_{i}{\bigg( -f'(x_t^i)\log(\frac{h(x_t^i)}{1 - h(x_t^i)}) \bigg)}\bigg|, \\
    \end{split}
    \end{equation}
    where $f'(x) = f_\mathcal{T}(x) - f_\mathcal{S}(x)$. 
    For the optimal hypothesis $h^*$, it is assumed that the empirical error in the source domain is small enough, $i.e.$, $\mathcal{R}_\mathcal{S}(h^*, f_\mathcal{S}) \rightarrow 0$, and the discriminability of $h^*$ in $\mathcal{T}$ is smaller than that in $S$, as
    \begin{equation}
    \label{eq:bound2}
    \begin{split}
        |\log(\frac{h^*(x_t^i)}{1 - h^*(x_t^i)})| &\leq  |\log(\frac{h^*(x_s^i)}{1 - h^*(x_s^i)})|.
    \end{split}
    \end{equation}
    We therefore have
    \begin{equation}
    \label{eq:bound3}
    \begin{split}
        -f'(x_t^i)\log(\frac{h^*(x_t^i)}{1 - h^*(x_t^i)}) &\leq  -f'(x_s^i)\log(\frac{h^*(x_s^i)}{1 - h^*(x_s^i)}).
    \end{split}
    \end{equation}
    Substituting Eq.\ \ref{eq:bound1} to Eq.\ \ref{eq:bound3}, we have
    \begin{equation}
    \label{eq:bound4}
    \begin{split}
        |\mathcal{R}_\mathcal{T}(h^*, f_\mathcal{T}) - \mathcal{R}_\mathcal{T}(h^*, f_\mathcal{S})| &\leq |\mathcal{R}_\mathcal{S}(h^*, f_\mathcal{T}) - \mathcal{R}_\mathcal{S}(h^*, f_\mathcal{S})|.
    \end{split}
    \end{equation}
    According to Eq.\ \ref{eq:bound}, the error bound of hypothesis $h^*$ in the target domain is concluded as
    \begin{equation}
    \label{eq:bound*}
    \begin{split}
        \mathcal{R}_\mathcal{T}(h^*, f_{\mathcal{T}}) 
        &\leq \mathcal{R}_\mathcal{T}(h^*, f_\mathcal{S}) + |\mathcal{R}_\mathcal{S}(h^*, f_{\mathcal{T}}) - \mathcal{R}_\mathcal{S}(h^*, f_{\mathcal{S}})| \\
        &\leq \mathcal{R}_\mathcal{T}(h^*, f_\mathcal{S}) + \mathcal{R}_\mathcal{S}(h^*, f_{\mathcal{T}}).
    \end{split}
    \end{equation}
    
    \textbf{Domain Contrast (DC) Loss}. 
    To quantify $\mathcal{R}_\mathcal{S}(h^*, f_\mathcal{T})$, $h^*$ is supposed to be a nearest neighbor classifier. The probability that a source domain sample $x_s^i$ has the same class label with it's neighbors in the target domains is calculated as their similarity $\text{sim}(x_s^i, x_{t}^i)$, where $\text{sim}(u,v)=u^{\top} v/||u||||v||$ defines the $cosine$ similarity between two samples. Combining the probabilities to a softmax function, the nearest neighbor classifier $h^*$ is defined as
    \begin{equation}
    \label{eq:softmax}
        h^*(x_t^i) = \frac{ \sum_j f_\mathcal{S}(x_s^j) \exp( \text{sim}(x_t^i, x_s^j) )}{\sum_{j}{\exp( \text{sim}(x_t^i, x_s^j))}},
    \end{equation}
    where $x_{t}^i$ is the $i^{th}$ sample transferred from the source to the target domain. Accordingly, minimization of the empirical error of a source model in the target domain, $\mathcal{R}_\mathcal{T}(h^*, f_\mathcal{S})$, can be implemented by minimizing
    \begin{equation}
    \label{eq:rt}
        \begin{split}
        \mathcal{R}_\mathcal{T}(h^*, f_\mathcal{S}) 
        &= \frac{1}{N} \sum_{i}{-\log\bigg( \frac{ \sum_j I(x_t^i, x_s^j) \exp( \text{sim}(x_t^i, x_s^j) )}{\sum_{j}{\exp( \text{sim}(x_t^i, x_s^j))}} \bigg)} \\
        &\leq \frac{1}{N} \sum_{i}{-\log\bigg( \frac{ \exp( \text{sim}(x_t^i, x_s^i) )}{\sum_{j}{\exp( \text{sim}(x_t^i, x_s^j))} )} \bigg)},
    \end{split}
    \end{equation}
    where $I(x_1, x_2) = 1 - |f_\mathcal{S}(x_1) - f_\mathcal{S}(x_2)|$. $N$ denotes the number of samples. Correspondingly, we approximately quantify the $\mathcal{R}_\mathcal{S}(h^*, f_\mathcal{T})$ as
    \begin{equation}
    \label{eq:rs}
        \begin{split}
        \mathcal{R}_\mathcal{S}(h^*, f_\mathcal{T}) 
        &\leq \frac{1}{N} \sum_{i}{-\log\bigg( \frac{ \exp( \text{sim}(x_s^i, x_t^i) )}{\sum_{j}{\exp( \text{sim}(x_s^i, x_t^j))} )} \bigg)}.
    \end{split}
    \end{equation}
    
    According Eqs.\ \ref{eq:bound4}, \ref{eq:rt} and \ref{eq:rs}, minimizing the error bound defined by Eq.\ \ref{eq:bound} can be fulfilled by optimizing network parameter $\theta$ to minimize
    \begin{equation}
    \label{eq:dc_loss}
    \begin{split}
        \mathcal{L}_C(\mathcal{S}, \mathcal{T})
        = -\frac{1}{N} \sum_{i}{\log\bigg( \frac{ \exp( \text{sim}(x_t^i, x_s^i) )}{\sum_{j}{\exp( \text{sim}(x_t^i, x_s^j))} )} \bigg)}
        - \frac{1}{N} \sum_{i}{\log\bigg( \frac{ \exp( \text{sim}(x_s^i, x_t^i) )}{\sum_{j}{\exp( \text{sim}(x_s^i, x_t^j))} )} \bigg)},
    \end{split}
    \end{equation}
    which is referred to as the DC loss.
     
     \textbf{Advantages of DC Loss}. The DC loss is derived from the perspective of error bound minimization, 
    while reflecting the similarity and dis-similarity (contrast) between samples across the domains. Minimizing DC loss drives learning feature representations that capture information shared between source and target domains but that are discriminative $i.e.$, different samples/instances in the two domains have small similarities ($i.e.$, large $cosine$ distances), Fig.\ \ref{fig:dc}. 
    
    It is known that during object detection the positive-negative class imbalance is an important issue. Such an issue has been widely explored in supervised detection, but unfortunately ignored by the domain adaptive detection, which could deteriorate the discriminability of transferred detectors. The nominator and denominator of the DC loss naturally incorporate sampling imbalance, which, during transfer, facilities alleviating class imbalance in the target domain. This is an advantage of DC loss compared to the Triplet Loss~\cite{TripletLoss2019}. 

\subsection{Domain Adaptive Detection}
 
     To implement domain adaptive detection, a base detector based on the deep network is first trained using the annotated data $\{x^n_s, y^n_s\}_{n=1}^{N^s}$ in the source domain by minimizing the detection loss $\mathcal{L}_{D}(\theta)$. The trained base detector is then transferred to the target domain by fine-tuning the network and minimizing the image- and region-level DC loss $\mathcal{L}^I_{C,\mathcal{S}}(\theta)$ and $\mathcal{L}^R_{C,\mathcal{S}}(\theta)$. In turn, the detector is transferred to the source domain by minimizing the image-level DC loss $\mathcal{L}^I_{C,\mathcal{T}}(\theta)$ and fine-tuning the detector using $\mathcal{L}_{\mathcal{T}}^{R}(\theta)$, which is the pseudo ground-truth loss.
    
\begin{figure}[t]
		\centering
		\includegraphics[width = 0.99\linewidth]{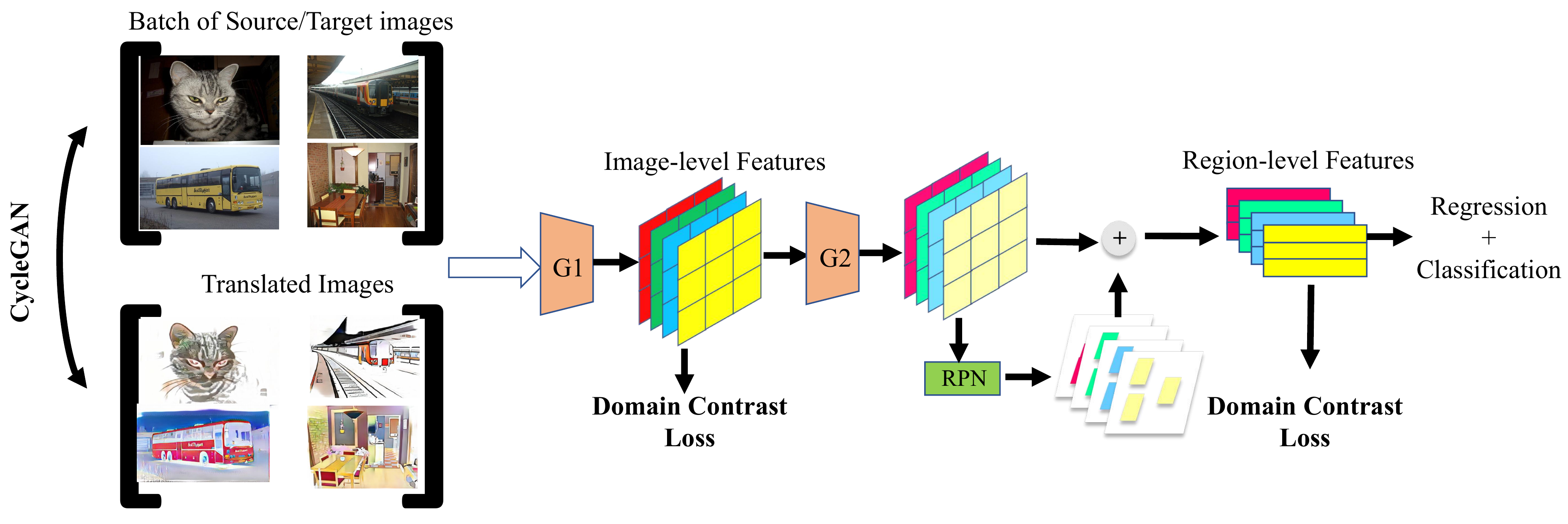}
		\caption{Illustration of the domain adaptive objector based upon CycleGAN and the proposed DC loss. CycleGAN is adopted to translate an image from the $\mathcal{S}/\mathcal{T}$ to the $\mathcal{T}/\mathcal{S}$ domain. $G1$ and $G2$ refer to convolutional layers. The detection network is first trained within $\mathcal{S}$ by minimizing the training loss $\mathcal{L}_{D}(\theta)$. The network then is transferred across domains by progressively minimizing the DC loss including $\mathcal{L}_{C,\mathcal{S}}^{I}(\theta)$, $\mathcal{L}_{C,\mathcal{S}}^{R}(\theta)$, and $\mathcal{L}_{C,\mathcal{T}}^{I}(\theta)$ and the pseudo ground-truth loss $\mathcal{L}_{\mathcal{T}}^{R}(\theta)$.}
		\label{fig:flowchart}
	\end{figure}

    \textbf{Base Detector}. The Faster R-CNN~\cite{ren2015faster} is employed as the base detector, which consists of three stages: convolutional feature extraction, region proposal generation (RPN) and bounding box prediction, Fig.\ \ref{fig:flowchart}. Each input image is represented as image-level features and RPN generates object region proposals, of which region-level features are extracted by ROI-pooling. With region-level features, the category labels and bounding boxes are predicted by classification and regression subnets. Detection loss $\mathcal{L}_{D}(\theta)$ is composed of the loss of the RPN and the loss of the subnets.
    
    \textbf{$\mathcal{S}\rightarrow\mathcal{T}$ Transfer}.  
    To transfer the Faster RCNN detector parameterized by $\theta$, we first translate each sample image $x_s^i$ from $\mathcal{S}$ to $\mathcal{T}$ using the CycleGAN method~\cite{CycleGAN2017}, which learns to translate an image from a source domain to a target domain given unpaired/paired examples. After iterative adversarial generation, the translated source images have similar styles with other images in the target domain.
    
    Denote the features of a translated sample as $x_{s\rightarrow t}^i(I;\theta)$. For a batch of samples in $\mathcal{S}$ and their translated counterparts in $\mathcal{T}$, we construct a positive sample pair $(x_s^i(I;\theta),x_{s\rightarrow t}^i(I;\theta))$ and $N-1$ negative sample pairs $(x_s^j(I;\theta),x_{s\rightarrow t}^i(I;\theta))$.
    Given positive negative sample pairs, $\mathcal{S}\rightarrow\mathcal{T}$ transfer is implemented by fine-tuning the network parameter to minimize DC loss. By introducing a temperature parameter $\tau$ to Eq.\ \ref{eq:dc_loss}, the image-level $\mathcal{S}\rightarrow\mathcal{T}$ DC loss is defined as
  \begin{equation}
  \begin{split}
    \label{eq:dc_st}
    \mathcal{L}_{C,\mathcal{S}}^{I}(\theta) 
    = &-\frac{1}{N} \sum_{i}{\log\bigg( \frac{ \exp( \text{sim}(x_{s\rightarrow t}^i(I;\theta), x_s^i(I;\theta))/\tau )}{\sum_{j}{\exp( \text{sim}(x_{s\rightarrow t}^i(I;\theta), x_s^j(I;\theta))/\tau)} )} \bigg)} \\
    &-\frac{1}{N} \sum_{i}{\log\bigg( \frac{ \exp( \text{sim}(x_s^i(I;\theta), x_{s\rightarrow t}^i(I;\theta))/\tau )}{\sum_{j}{\exp( \text{sim}(x_s^i(I;\theta), x_{s\rightarrow t}^j(I;\theta))/\tau)} )} \bigg)},
 \end{split}
 \end{equation}
 where $x^i(I;\theta)$ denotes the features in the last convolutional layer for a sample image. 
 
 In each image, we randomly select at the most two ground-truth bounding boxes for transfer. The features of selected proposals in an image are concatenated to a long vector, denoted as $x^i(R;\theta)$. The region-level contrast loss $\mathcal{L}_{C,\mathcal{S}}^{R}(\theta)$ can be calculated by replacing the image-level features $x^i(I;\theta)$ with the region-level features $x^i(R;\theta)$. 
   
    \textbf{$\mathcal{T}\rightarrow\mathcal{S}$ Transfer}.  
    In a similar way the image-level $\mathcal{T}\rightarrow\mathcal{S}$ DC loss is defined as
    \begin{equation}
    \begin{split}
    \label{eq:dc_ts}
    \mathcal{L}_{C,\mathcal{T}}^{I}(\theta) 
    = &-\frac{1}{N} \sum_{i}{\log\bigg( \frac{ \exp( \text{sim}(x_t^i(I;\theta), x_{t\rightarrow s}^i(I;\theta)/\tau )}{\sum_{j}{\exp( \text{sim}(x_t^i(I;\theta), x_{t\rightarrow s}^j(I;\theta)/\tau)} )} \bigg)}\\
    &-\frac{1}{N} \sum_{i}{\log\bigg( \frac{ \exp( \text{sim}(x_{t\rightarrow s}^i(I;\theta), x_t^i(I;\theta)/\tau )}{\sum_{j}{\exp( \text{sim}(x_{t\rightarrow s}^i(I;\theta), x_t^j(I;\theta)/\tau)} )} \bigg)},
    \end{split}
    \end{equation}
 
  As there is no ground-truth object annotated in the target domain, the region-level transfer can not be directly applied. We thereby first translate all the training image from $\mathcal{T}$ to $\mathcal{S}$ using the CycleGAN method. We then use the detector trained in $\mathcal{S}$ to detect high-scored regions in the translated images as pseudo ground-truth objects. With the pseudo ground-truth objects, region-level $\mathcal{T}\rightarrow\mathcal{S}$ transfer is carried out by fine-turning the detector to minimize the detection loss $\mathcal{L}_{\mathcal{T}}^{R}(\theta)$ for pseudo ground-truth objects.


\section{Experiment}
    We analyzed the proposed domain adaptation method by training a detector in the PASCAL VOC dataset and transferring~\cite{Pascal2010} it to the Clipart dataset~\cite{Clipart2018} for performance evaluation. We also transferred detectors across domains, Pascal VOC $\rightarrow$ Comic2K and PASCAL $\rightarrow$ WaterColor2K~\cite{Clipart2018}, where large domain divergence exists.

    \subsection{Experimental Setting}
    Faster R-CNN with the ResNet101~\cite{ResNet2016} backbone pre-trained on the ImageNet~\cite{ImageNet2009} was employed as the base detector. While training the domain adaptive detector, the inputs were a mini-batch of images pairs, including $N$ (batch size) annotated images from the source domain and $N$ unannotated images from the target domain. We set the shorter side of the image to 600 pixels following the setting in of Faster RCNN~\cite{ren2015faster}. The detection network (detector) was trained with a learning rate of 0.001 in the first 5 epochs and decreased to 0.0001 in the following 2 epochs. The iteration number of each epoch is calculated by the sample number divided by the batch size. 
    The mean Average Precision(mAP) for all object categories in the dataset was used as the evaluation metric.
    
    The DC loss is plug-and-play, which means it operates simply by fine-tuning the detector trained in the source domain. 
    For each DC loss, the detector was fine-tuned by 5 epochs with a learning rate determined by experiments (Fig.\ \ref{fig:paras}). In this way, there is no regularization factor required to balance the importance of each loss terms defined in Sec. 3.2, which simplifies the parameter settings.

  	\begin{figure}[t]
		\centering
		\begin{minipage}[t]{0.45\linewidth}
		\includegraphics[width = 1\linewidth]{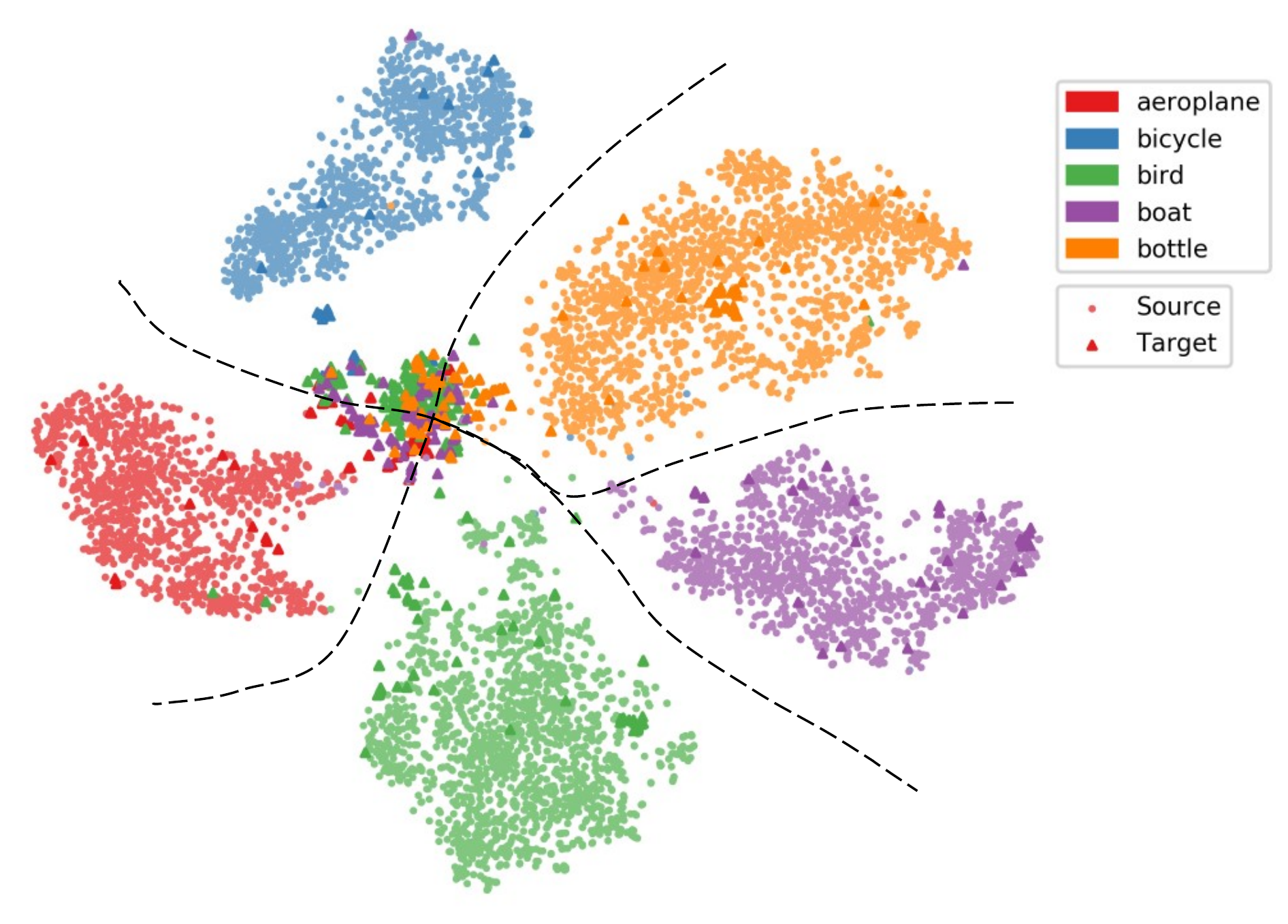}
		\end{minipage}
		\begin{minipage}[t]{0.45\linewidth}
		\includegraphics[width = 1\linewidth]{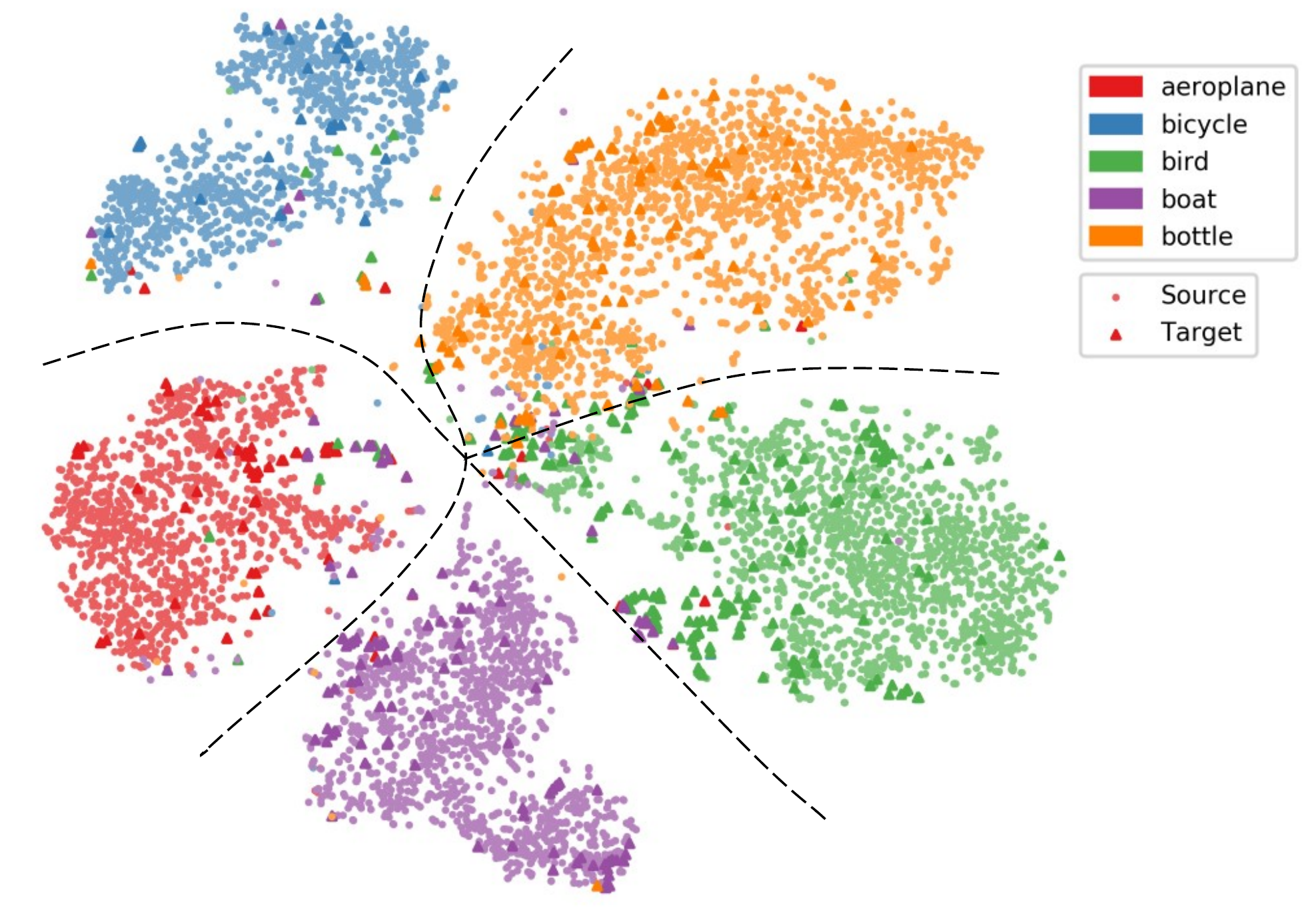}
		\end{minipage}
		\caption{t-SNE of samples before (left) and after (right) domain adaptation. (Best viewed in color) }
		\label{fig:tSNE}
	\vspace{-0.5cm}
	\end{figure}
	
	\subsection{Model Effect} 
	
	In Fig.\ \ref{fig:tSNE}, we compared the distributions of target samples before and after domain adaptation. Before adaptation, the target domain samples tends to be concentrated together and are difficult to discriminate. After domain adaptation using DC, the distribution of the target domain samples were well aligned with that of the source domain samples. At the same time, target samples can be well discriminated, which demonstrated the effect of the proposed DC method.
	
	\begin{table*}[!t]
		\renewcommand{\arraystretch}{1.3}
		\newcommand{\tabincell}[2]{\begin{tabular}{@{}#1@{}}#2\end{tabular}}
		\caption{Ablation study of detection performance (mAP\%) and comparison with the state-of-the-arts when transferring a detector trained within PASCAL VOC to Clipart.}
		\label{tab:abl}
		\centering \scriptsize
		\begin{tabular}{
		    p{2.25cm}@{}
		    @{}m{0.54cm}<{\centering}
		    @{}m{0.54cm}<{\centering}
		    @{}m{0.54cm}<{\centering}
		    @{}m{0.54cm}<{\centering}
		    @{}m{0.54cm}<{\centering}@{}
			@{}m{0.54cm}<{\centering}
			@{}m{0.52cm}<{\centering}
			@{}m{0.52cm}<{\centering}
			@{}m{0.56cm}<{\centering}
			@{}m{0.52cm}<{\centering}@{}
			@{}m{0.56cm}<{\centering}
			@{}m{0.56cm}<{\centering}
			@{}m{0.56cm}<{\centering}
			@{}m{0.60cm}<{\centering}
			@{}m{0.60cm}<{\centering}@{}
			@{}m{0.60cm}<{\centering}
			@{}m{0.60cm}<{\centering}
			@{}m{0.60cm}<{\centering}
			@{}m{0.54cm}<{\centering}
			@{}m{0.54cm}<{\centering}
			|@{}m{0.70cm}<{\centering}@{}
		}
		\hline
		Method & aero & bike & bird & boat & bott. & bus & car & cat & chair & cow & table & dog & hrs & mbike & pers. & plant & sheep & sofa & train & tv & mAP  \\
		WST-BSR\cite{WSTBSR2019} & 35.6 & 52.5 & 24.3 & 23.0 & 20.0 & 43.9 & 32.8 & 10.7 & 30.6 & 11.7 & 13.8 & 6.0 & 36.8 & 45.9 & 48.7 & 41.9 & 16.5 & 7.3 & 22.9 & 32.0 & 27.8\\
		SWDA\cite{SWDA2019} & 26.2 & 48.5 & 32.6 & 33.7 & 38.5 & 54.3 & 37.1 & 18.6 & 34.8 & \bf58.3 & 17.0 & 12.5 & 33.8 & 65.5 & 61.6 & 52.0 & 9.3 & 24.9 & \bf54.1 & 49.1 & 38.1\\
		ICR-CCR~\cite{ICR-CCR2020}&28.7& 55.3& \bf31.8& 26.0& 40.1& \bf63.6& 36.6& 9.4& 38.7&49.3& 17.6& 14.1& 33.3& 74.3& 61.3& 46.3& 22.3& 24.3& 49.1& 44.3& 38.3\\
		HTCN\cite{Chen2020Transferability} &33.6 & 58.9 & 34.0 & 23.4 & 45.6 &57.0 & 39.8 & 12.0 & 39.7 & 51.3 & 21.1 & 20.1 & 39.1 & 72.8 & 63.0 & 43.1 & 19.3 & 30.1 & 50.2 & 51.8 & 40.3\\
		DM\cite{kim2019diversify} &25.8 &\bf63.2 & 24.5 &\bf42.4 & 47.9 & 43.1 & 37.5 & 9.1 & \bf47.0 & 46.7 & 26.8 & \bf24.9 & \bf48.1 & 78.7 & 63.0 & 45.0 & 21.3 & 36.1 & 52.3 & \bf53.4 & 41.8\\
		\hline
		Baseline & 35.6 & 52.5 & 24.3 & 23.0 & 20.0 & 43.9 & 32.8 & 10.7 & 30.6 & 11.7 & 13.8 & 6.0 & 36.8 & 45.9 & 48.7 & 41.9 & 16.5 & 7.3 & 22.9 & 32.0 & 27.8\\
		DC$^{I}_{\mathcal{S\rightarrow T}}$ & 44.0& 49.4& 34.9& 34.0& 40.1& 52.4& 42.2& 11.5& 38.4& 37.1& 30.4&15.6& 34.0& 84.6& 58.5& 50.2& 14.4& 24.5& 35.6& 42.3& 38.7\\
		DC$^{R}_{\mathcal{S\rightarrow T}}$&29.4& 53.2& 27.4& 26.4& 45.2& 51.5& 41.0& 5.2& 35.8& 	36.5& 22.3& 9.8& 31.7& 79.1& 51.6& 42.3& 12.5& 25.3& 43.6& 41.2& 35.5\\
		DC$^{I,R}_{\mathcal{S\rightarrow T}}$&40.3&58.7& 33.0& 31.9& \bf49.3& 51.9& 47.2& 6.5& 36.8& 38.3& \bf32.1& 16.7& 32.2& 85.3& 57.9& 48.0& 15.0& 25.9& 46.3& 44.2& 39.9\\
		DC$^{I}_{\mathcal{T\rightarrow S}}$ &36.0& 53.1& 29.9& 24.6& 40.1& 51.0& 33.9& 7.5& 39.2&  23.8&  23.2& 11.5& 	31.4& 59.7& 41.9& 49.2& 10.9& \bf30.4& 47.9& 41.1& 34.3\\
		DC$^{I,R}_{\mathcal{S\rightarrow T}}$-DC$^{I}_{\mathcal{T\rightarrow S}}$ &45.2& 55.9& 33.8& 32.8& 49.2& 52.2& 48.2& 9.4& 37.6& 38.7& 31.8& 16.6& 34.9& \bf87.3& 60.3& 50.2& 15.8& 27.4& 45.5& 47.9& 41.0\\
		DC$^{I,R}_{\mathcal{S\rightarrow T}}$- DC$^{I,R}_{\mathcal{T\rightarrow S}}$&\bf47.1& 53.2& \bf38.8& 37.0& 46.6& 45.8& \bf52.6& \bf14.5& 39.1& 48.4& 31.7& 23.7& 34.9& 87.0& \bf67.8& \bf54.0& \bf22.8& 23.8 & 44.9& 51.0& \bf43.2\\
		\hline
		Oracle & 33.3& 47.6& 43.1& 38.0& 24.5& 82.0& 57.4& 22.9& 48.4& 49.2& 37.9& 46.4& 41.1& 54.0& 73.7& 39.5& 36.7& 19.1& 53.2& 52.9 &45.0\\
		\hline
		
		\end{tabular}
	\end{table*}
    
	\textbf{Ablation Study.} In Table\ \ref{tab:abl}, the effect of domain contrast was validated by performing $\mathcal{S}\rightarrow \mathcal{T}$ and $\mathcal{T}\rightarrow \mathcal{S}$ transfer step-by-step. Using solely the image-level $\mathcal{S}\rightarrow \mathcal{T}$ transfer (DC$^{I}_{\mathcal{S\rightarrow T}}$) the mAP is improved by 10.9\% (38.7\% vs. 27.8\%) which validated the effectiveness of our approach at reducing cross-domain divergence while improving detector discriminability. The region-level $\mathcal{S}\rightarrow \mathcal{T}$ transfer (DC$^{R}_{\mathcal{S\rightarrow T}}$) improved the mAP by 7.7\%. Combining the image-level and region-level $\mathcal{S}\rightarrow \mathcal{T}$ transfer (DC$^{I,R}_{\mathcal{S\rightarrow T}}$) improved the mAP by 12.1\% (39.9\% vs. 27.8\%).
	
    \begin{figure}[t]
		\centering
		\includegraphics[width = 0.8\linewidth]{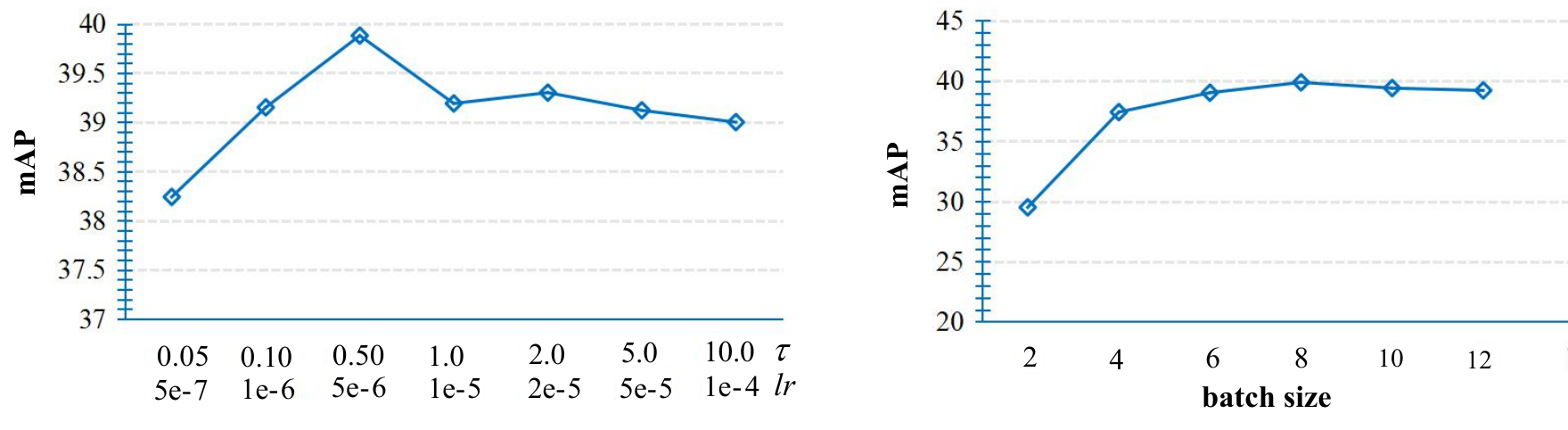}
		\caption{Ablation of parameter $\tau$, learning rate ($lr$), and batch size.}
    	\label{fig:paras}
    \vspace{-0.5cm}
	\end{figure}
	
	After performing image-level $\mathcal{T}\rightarrow \mathcal{S}$ transfer (DC$^{I,R}_{\mathcal{S\rightarrow T}}$-DC$^{I}_{\mathcal{T\rightarrow S}}$), the mAP is further improved by 1.1\% (41.0\% vs. 39.9\%). Using the pseudo objects in the target domain to perform region-level $\mathcal{T}\rightarrow \mathcal{S}$ transfer was also effective, improving the mAP by 2.2\% (43.2\% vs. 41.0\%). To reduce false positives, the threshold for pseudo object detection was set to be 0.95. Without whistle and bells, our method outperformed the state-of-the-art by 2.4\% (43.2\% vs. 41.8\%), which was a significant margin considering the challenging task. We reported the ``Oracle" result by training a Faster RCNN detector using the images within target domain but with the ground truth annotations, which is a reference for the performance upper-bound. 
	
    \begin{wrapfigure}{r}{5.5cm}
    \begin{center}
    \includegraphics[height=0.55\linewidth]{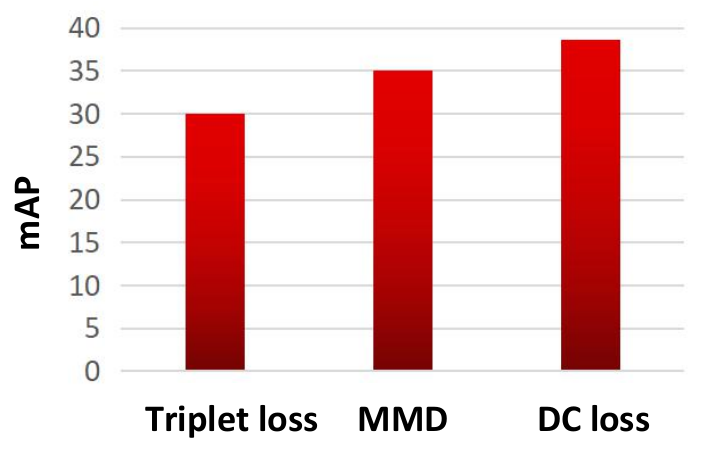}
    \end{center}
    \caption{Comparison of mAPs of DC loss, Triplet Loss and MMD by performing image-level transfer from Pascal VOC to Clipart.}
    \label{fig:hist}
    \end{wrapfigure}

	\textbf{Parameter Setting.} In Fig.\ \ref{fig:paras}, parameters $\tau$ and batch size were analyzed for DC loss. For each $\tau$, we determined the best learning rate by a linear search strategy. With a learning rate $5e$-6 and $\tau=0.5$ the best performance was achieved. The optimal $\tau$ selection improved 1.75\% mAP, showing the importance of the temperature parameter. In contrast, the performance was not very sensitive to the batch size, $e.g.$, using a large batch size 8 slightly improved the mAP. This is different from unsupervised contrastive learning which often relies on large batch sizes~\cite{CMC2019,Sim2020}.  
	
    \textbf{Imbalance Issue.} In Fig.\ \ref{fig:hist}, we present the advantage of DC loss over the Triplet loss~\cite{TripletLoss2019}, which minimizes the intra-class distance of each positive sample pair and maximizes the inter-class distances of each positive-negative sample pair as $\sum_j \max(||x_s^i-x_{s\rightarrow t}^i||^2_2-||x_s^i-x_{s\rightarrow t}^j||^2_2+0.5,0)$. Triplet Loss reported much lower performance as it pursued maximizing the domain similarity and minimizing the similarity of a single pair of sample in a mini-batch but ignored the class imbalance issue in object detection. Such an imbalance issue was also ignored by the MMD method~\cite{LargeNorm2019}. 
    

    \subsection{Performance and Comparison}

    \begin{table*}[!t]
      \renewcommand{\arraystretch}{1.3}
		\newcommand{\tabincell}[2]{\begin{tabular}{@{}#1@{}}#2\end{tabular}}
		\caption{Performance comparison when transferring detectors from Pascal VOC to Comic2k.}
		\label{tab:Comic2K}
		\centering \scriptsize
		\begin{tabular}{
		    p{2.25cm}@{}|
		    @{}m{3.00cm}<{\centering}|
		    @{}m{0.55cm}<{\centering}
		    @{}m{0.55cm}<{\centering}
		    @{}m{0.55cm}<{\centering}
		    @{}m{0.55cm}<{\centering}@{}
			@{}m{0.55cm}<{\centering}
			@{}m{0.80cm}<{\centering}|
			@{}m{0.55cm}<{\centering}
		}
		\hline
		Method & Base Detector & bike & bird &car	&cat	&dog	&person	&mAP  \\
		DT\cite{Clipart2018}      &SSD+VGG16         & 43.6	&13.6	&30.2	&16.0	&26.9	&48.3	&29.8  \\
		WST-BSR\cite{WSTBSR2019}  &SSD+VGG16         & 50.6	&13.6 &31.0	&7.5 &16.4 &41.4 &26.9 \\
		DM\cite{kim2019diversify} & Faster RCNN+VGG16 & - & - &-	&-	&-	&-	&34.5  \\
		\hline
		Baseline  & Faster RCNN+VGG16 &38.5	&10.5	&14.7	&15.1	&15.2	&29.8	&20.6\\
                  & Faster RCNN+ResNet101 &30.7	&13.8 &24.2	&13.8	&14.8	&32.7	&21.7\\
		DC (ours)      & Faster RCNN+VGG16 &\bf52.7	&17.4	&\bf43.4	&23.3	&25.9	&58.7	&36.9\\
                      & Faster RCNN+ResNet101 &51.9	&\bf23.9	&36.7	&\bf27.1	&\bf31.5	&\bf61.0	&\bf38.7\\
 		\hline
	
		\end{tabular}
		\vspace{-0.2cm}
	\end{table*}
    
     \begin{table*}[!t]
      \renewcommand{\arraystretch}{1.3}
		\newcommand{\tabincell}[2]{\begin{tabular}{@{}#1@{}}#2\end{tabular}}
		\caption{Performance comparison when transferring detectors from Pascal VOC to WaterColor2K.}
		\label{tab:WaterColor2K}
		\centering \scriptsize
		\begin{tabular}{
		    p{2.25cm}@{}|
		    @{}m{3.00cm}<{\centering}|
		    @{}m{0.55cm}<{\centering}
		    @{}m{0.55cm}<{\centering}
		    @{}m{0.55cm}<{\centering}
		    @{}m{0.55cm}<{\centering}@{}
			@{}m{0.55cm}<{\centering}
			@{}m{0.80cm}<{\centering}|
			@{}m{0.55cm}<{\centering}
		}
		\hline
		Method & Base Detector & bike & bird &car	&cat	&dog	&person	&mAP  \\
		DT\cite{Clipart2018}      &SSD+VGG16         & \bf82.8	&47.0	&40.2	&34.6	&35.3	&62.5	&50.4  \\
		WST-BSR\cite{WSTBSR2019}  &SSD+VGG16         & 75.6	&45.8 &\bf49.3	&34.1 &30.1 &64.1 & 49.9 \\
		DM\cite{kim2019diversify} & Faster RCNN+VGG16 & - & - &-	&-	&-	&-	&52.0  \\
		SWDA\cite{SWDA2019}       & Faster RCNN+ResNet101 & 82.3 &\bf55.9 &46.5	&32.7	&\bf35.5	&66.7	&\bf53.3  \\
		\hline
		Baseline   & Faster RCNN+ResNet101 &30.7	&13.8 &24.2	&13.8	&14.8	&32.7	&21.7\\
		DC (ours)  & Faster RCNN+ResNet101 &76.7	&53.2	&45.3	&\bf41.6 &\bf35.5 &\bf70.0 & \bf53.7\\
 		\hline
 		Oracle    & Faster RCNN+ResNet101 &49.8	&50.6	&40.2	&38.9 &53.3 &69.4 & 50.4\\
 		\hline
	
		\end{tabular}
		\vspace{-0.2cm}
	\end{table*}
	
\begin{wrapfigure}{r}{8cm}
    \vspace{-0.8cm}
    \begin{center}
    \includegraphics[height=0.6\linewidth]{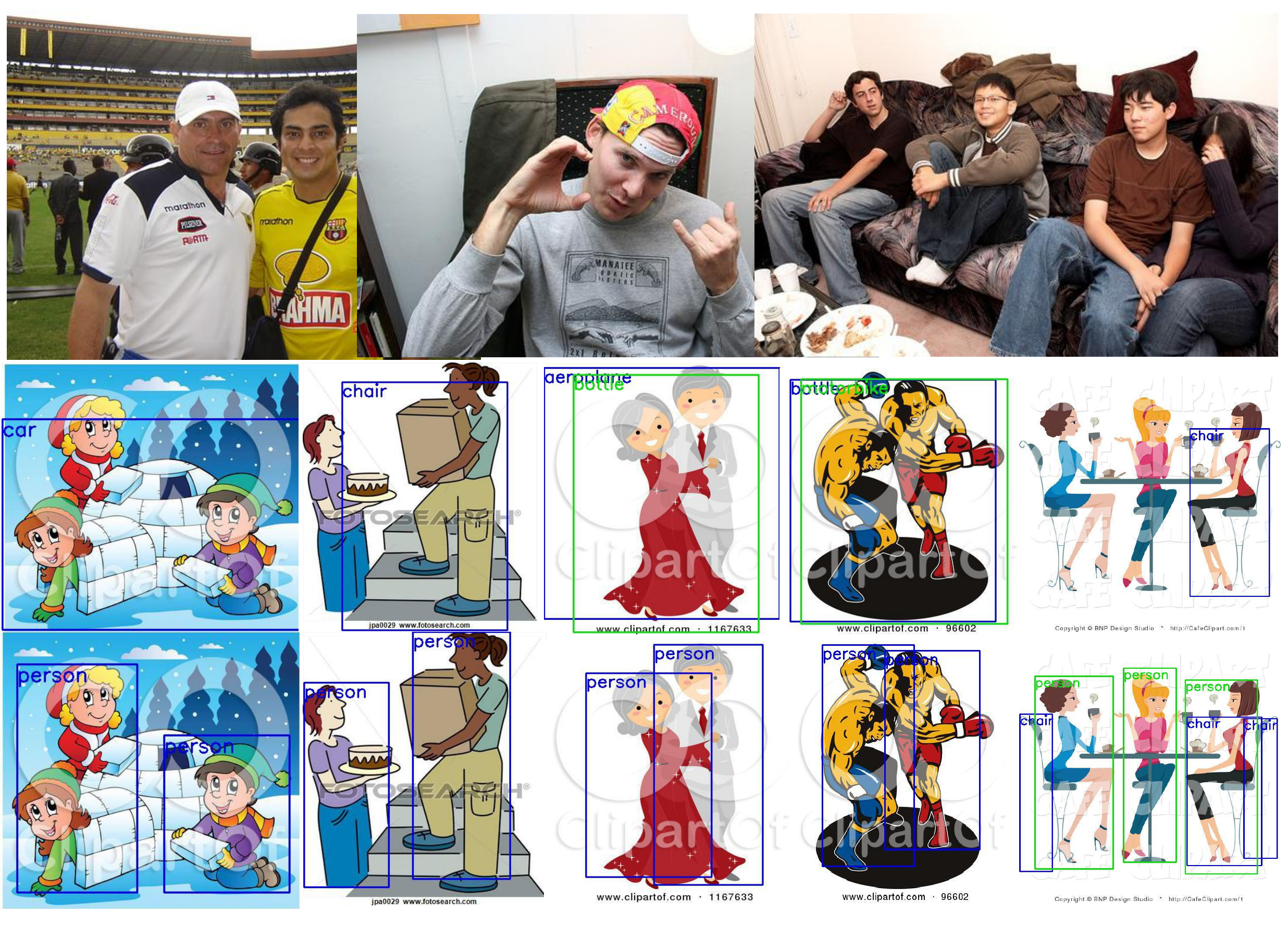}
    \end{center}
    \caption{Detection examples from the Clipart dataset. Images from the source domain are in the first row. The results of the baseline method is in the second row and our detection results are in the last row. (Best viewed in color)}
    \label{fig:examples}
    \vspace{-1.2cm}
    \end{wrapfigure}

    In Table\ \ref{tab:Comic2K}, we evaluated the proposed method and compared it with state-of-the-art methods when transferring detectors from Pascal VOC to Comic2k~\cite{Clipart2018}. The Comic2k dataset includes 2,000 comic images, 1,000 for training and the other 1,000 for test. It has 6 object classes which also exist in Pascal VOC. As the images in Comic2k are unrealistic images, Fig.\ \ref{fig:examples}, the domain divergence between the source (VOC) and the target domain (Comic2k) was predictably large. In this scenario, the proposed DC method outperformed the state-of-the-art method  WST-BST~\cite{WSTBSR2019} by 10.0\% (36.9\% vs. 26.9\%) and DM~\cite{kim2019diversify} by 2.4\% (36.9\% vs. 34.5\%). Using the ResNet-101 backbone further boosted the performance to 38.7\%. 
    
    In Table\ \ref{tab:WaterColor2K}, our method also outperformed the state-of-the-arts even 
    the room for improvement was very small. The detection examples in Fig.\ \ref{fig:examples} demonstrated the effectiveness of our method.

\section{Conclusion}

    We presented Domain Contrast (DC), a simple yet effective approach to train  domain adaptive detectors. This DC method is theoretically plausible because it was deduced from the perspective of error bound minimization about transfer learning. This DC method is conceptually simple and it can simultaneously guarantee the transferability of detectors while preserving the discriminability of transferred detectors by minimizing DC loss. DC significantly boosted the performance of domain adaptive detectors and improved the state-of-the-art on benchmark datasets of large domain divergence. This is the first time that the contrastive learning method has been applied to the domain adaptation problem and therefore provides a fresh insight to transfer learning. 
    
\newpage    
\section*{Broader Impact}   
\textbf{Benefit:} The transfer learning community may benefit from this research. The proposed Domain Contrast method provides a fresh insight to transfer learning, which appears to have reached a plateau due to increasingly complex methodologies in recent years. With the code publicly available (please refer to the supplementary material), our research will also benefit the industry by improving the generalization capability of detectors trained within a label-rich domain ($i.e.$, publicly annotated datasets) to an unlabeled domain ($i.e.$, real-world scenarios).

\textbf{Disadvantage:} Our research may place researchers who continue using complex methodologies at a disadvantage. Our approach is simple, elegant, and effective, and could therefore be challenged by the status of quo of transfer learning.

\textbf{Consequence:} The failure of domain adaptation detection will not bring serious consequences, as the detectors trained in source domains can be the lower bound of the detection performance.

\textbf{Data Biases:} The proposed DC method DOES NOT leverage  biases in the data. Whereas, the domain adaptive method aims to overcome data biases across domains.
    
    \renewcommand\refname{References}
    \bibliographystyle{plainnat}
    \bibliography{main.bbl}

\end{document}